# The Visual Experience Dataset: Over 200 Recorded Hours of Integrated Eye Movement, Odometry, and Egocentric Video

Greene, M.R.[1,2], Balas, B.J.[3*], Lescroart, M.D.[4*], MacNeilage, P.R[4*], Hart, J.A.[2], Binaee, K.[5,4**], Hausamann, P.A.[6**], Mezile, R.[2**], Shankar, B.[4,7**], Sinnott, C.B.[4,8**], Capurro, K.[4***], Halow, S.[4***], Howe, H.[4***], Josyula, M.[2***], Li, A.[2], Mieses, A.[2***], Mohamed, A.[2***] Nudnou, I.[3***], Parkhill, E.[2***], Riley, P.[2***], Schmidt, B.[2***], Shinkle, M.W.[4***], Si, W.[2***], Szekely, B[4***], Torres, J.M.[2***], Weissmann, E.[2***].

1 - Barnard College, Columbia University
2 - Bates College
3 - North Dakota State University
4 - University of Nevada, Reno
5 - Magic Leap
6 - Technical University of Munich
7 - Unmanned Ground Systems
8 - Smith-Kettlewell Eye Research Institute

*, **, and *** indicate equal contribution and are listed in alphabetical order.

# Abstract

We introduce the Visual Experience Dataset (VEDB), a compilation of over 240 hours of egocentric video combined with gaze- and head-tracking data that offers an unprecedented view of the visual world as experienced by human observers. The dataset consists of 717 sessions, recorded by 58 observers ranging from 6-49 years old. This paper outlines the data collection, processing, and labeling protocols undertaken to ensure a representative sample and discusses the potential sources of error or bias within the dataset. The VEDB's potential applications are vast, including improving gaze tracking methodologies, assessing spatiotemporal image statistics, and refining deep neural networks for scene and activity recognition. The VEDB is accessible through established open science platforms and is intended to be a living dataset with plans for expansion and community contributions. It is released with an emphasis on ethical considerations, such as participant privacy and the mitigation of potential biases. By providing a dataset grounded in real-world experiences and accompanied by extensive metadata and supporting code, the authors invite the research community to utilize and contribute to the VEDB, facilitating a richer understanding of visual perception and behavior in naturalistic settings.

# Introduction and Motivation

Visual perception is shaped by experience in multiple ways. For example, the tuning properties of visual neurons reflect color and orientation statistics of natural visual environments (Lee et al., 2002; Olshausen & Field, 1996). Detection and recognition performance depends on the relative frequency of specific stimuli in an observer's experience (e.g., colors (Juricevic & Webster, 2009; Webster et al., 2007); faces (Meissner & Bheadsetham, 2001); and object configurations within scenes (Biederman et al., 1982; Davenport & Potter, 2004; Greene et al., 2015)). These findings support the hypothesis that the brain leverages the statistical redundancies in stimuli to enable efficient behavior (Attneave, 1954; Barlow, 2001). Thus, a complete characterization of the mechanisms of our visual system relies on a thorough understanding of the natural statistics of visual experience.

The basic statistics of photographs (contrast, edge orientations, luminance, color, and Fourier amplitude spectra) have been extensively studied (Geisler, 2008; Hansen et al., 2003; Harrison, 2022; Howe & Purves, 2004; Lee et al., 2002; Olshausen & Field, 1996; Tolhurst et al., 1992; Torralba & Oliva, 2003). Additional work has sought to characterize mid-level statistics such as texture, scale, and depth (Held et al., 2012; Long et al., 2016; Nishida, 2019; Portilla & Simoncelli, 2000; Ruderman, 1994; Sato et al., 2019; Su et al., 2013; Torralba & Oliva, 2002), as well as higher-order statistics, such as distributions of objects and their locations in the image plane (Greene, 2013; Zhou et al., 2018) and in depth (Adams et al., 2016; Song et al., 2015).

Despite these successes, several methodological limitations constrain the insights we can obtain from these studies. First, the vast majority of work on natural scene statistics is on static photographs, while visual experience unfolds over time. Although some studies of spatiotemporal scene statistics exist (Dong & Atick, 1995; DuTell et al., 2024; Rao & Ballard, 1998), we know comparatively little about the statistics of the motion experienced by the visual system, the extent to which this motion results from self-movement or the movement of other objects and surfaces, or the extent to which motion cues may contribute to perceptual invariants for recognition (Gibson, 1986). Second, many extant datasets are comparatively small or narrowly sample a small number of environments (e.g., Betsch et al., 2004). These datasets do not capture the full range of our visual experiences and thus may miss features of the natural world that affect perception. More troublingly, small datasets are particularly subject to bias (Tommasi et al., 2017; Torralba & Efros, 2011). That is, datasets capture unintended covariance related to the chosen semantic labels, as well as image-specific preferences such as viewpoint and lighting conditions. A related concern is that many datasets are sampled broadly from Internet images (Deng et al., 2009; Greene, 2013; Krishna et al., 2017; Lin et al., 2014; Xiao et al., 2014; Zhou et al., 2018). Internet-derived datasets have been critiqued for representing a narrow range of geographical locations (DeVries et al., 2019), viewpoints (Tseng et al., 2009), and socioeconomic circumstances (Tolia-Kelly, 2016); for containing harmful content such as pornography (Prabhu & Birhane, 2020); and for over-representing content that perpetuates race and gender stereotypes (Hirota et al., 2022; Kay et al., 2015; Otterbacher et al., 2017; Zhao et al., 2021). Finally, human vision is an active process. In order to fully characterize natural scene statistics, it is necessary to determine what observers fixate in the world rather than simply what falls in front of their head.

While several large video datasets have been developed by scientists in computer vision (see **Table 1**), none have been designed with these issues at the forefront. Many existing datasets provide dynamic views of a range of human actions, for example, but most are not taken from an egocentric perspective, and very few include the location of gaze or head (or camera) position. Thus, despite the impressive scope of some datasets (Grauman et al., 2022), it is unclear from these data, for example, where objects typically fall in the visual field, because little or no gaze data were collected. By contrast, several studies have also collected samples of mobile eye tracking data (e.g., (Hayhoe & Ballard, 2014; Kothari et al., 2020; Peterson et al., 2016; Sprague et al., 2015), but these have tended to focus on eye movements in particular circumstances, such as specific environments (Matthis et al., 2018), specific tasks such as making food or coffee (Fathi et al., 2011; Hayhoe & Ballard, 2005) or by examining gaze patterns to specific stimuli such as faces (Peterson et al., 2016). Other mobile eye tracking datasets have been collected for developing robust gaze tracking pipelines (Fuhl et al., 2021; Kothari et al., 2020). While these have provided useful insights into how gaze behavior varies across tasks, environments, and observers in constrained settings, these sets are comparatively small in scope and narrowly focused on the specific question they set out to answer. With a larger and more diverse dataset, we can examine many more general questions: How many faces, bodies, and animals do people regularly see? How many vertical surfaces do we see compared to oblique or horizontal surfaces? How many different types of objects do people typically encounter, and where in the visual field do these objects land? How do we move our heads and eyes during different types of activities? How does this affect our visual input? *What does the world, as experienced by acting human observers, really look like?*

In this work, we introduce the Visual Experience Dataset, a source of over 240 hours of egocentric video combined with gaze- and head tracking. This dataset was collected across two universities and funded by the National Science Foundation (1920896 to MRG). Following the recommendations of (Gebru et al., 2021), we have organized the manuscript to provide potential end-users with information on the composition, data collection, and data processing and labeling. We will describe current and possible uses for this dataset and detail the distribution of data and code that comprise the dataset, as well as our plans to maintain the dataset. Ultimately, we hope that by sharing this resource with the vision science community, progress can be made on several topics at the intersection of visual ecology, embodied cognition, and visual perception.

| Name | Ref. | Hours | Obs. | Places | Tasks | Ego? | Gaze? | IMU? |
|---|---|---|---|---|---|---|---|---|
| CMU Multimodal | (De la Torre Frade et al., 2009) | 54 | 43 | 1 | 5 cooking only | ✅ | ✖ | ✅ |
| MPI Cooking | (Rohrbach et al., 2012) | 8 | 12 | 1 | 65 cooking only | ✖ | ✖ | ✖ |
| Activities of Daily Living | (Pirsiavash & Ramanan, 2012) | 10 | 20 | 20 | 18 | ✅ | ✖ | ✖ |
| GTEA | (Fathi, Li, et al., 2012) | 28 | 32 | 1 | 86 cooking only | ✅ | ✅ | ✖ |
| Disney | (Fathi, Hodgins, et al., 2012) | 42 | 8 | 1 | 1 | ✅ | ✖ | ✖ |
| UT Ego | (Lee et al., 2012) | 17 | 4 | ? | ? | ✅ | ✖ | ✖ |
| UT | (Su & Grauman, 2016) | 14 | 9 | 3 | 3 | ✅ | ✖ | ✖ |
| Krishna Cam | (Singh et al., 2016) | 70 | 1 | ? | ? | ✅ | ✖ | ✖ |
| Something something | (Goyal et al., 2017) | 244 | 1133 | ? | 174 human-object pairs | ✅ | ✖ | ✖ |
| VLOG | (Fouhey et al., 2017) | 344 | 30,000+ | ? | YouTube samples | ✖ | ✖ | ✖ |
| THUMOS | (Idrees et al., 2017) | 430 | 10,697 | ? | YouTube samples | ✖ | ✖ | ✖ |
| Every moment counts | (Yeung et al., 2018) | 30 | ? | ? | YouTube samples | ✖ | ✖ | ✖ |
| Charades Ego | (Sigurdsson et al., 2018) | 71 | 112 | 15 indoor only | 157 granular | ✅ | ✖ | ✖ |
| You2me | (Ng et al., 2020) | 1.5 | 10 | 1 | 4 | ✅ | ✖ | ✖ |
| EPIC kitchens | (Damen et al., 2022) | 100 | 37 | 1 (45 kitchens) | 1 | ✅ | ✖ | ✖ |
| Ego4D | (Grauman et al., 2022) | 3670 | 931 | ? | 1772 verbs | ✅ | ⍻ | ⍻ |
| EgoCom | (Northcutt et al., 2023) | 39 | 34 | 1 | 1 | ✅ | ✖ | ✖ |
| Visual Experience Dataset | This work | 214 | 44 | 124 | 396 | ✅ | ✅ | ✅ |

**Table 1: Comparing related works. ✅indicates presence, √ indicates partial presence, ✖ indicates absence.**

# Dataset Composition

The Visual Experience Dataset is composed of 717 egocentric video sessions recorded by 56 individual observers ranging in age from 7 to 46 (22 female, 33 male, and one nonbinary). The Institutional Review Boards of Bates College, North Dakota State University, and University of Nevada Reno approved the recording protocols, and all participants (or their guardians, in the case of children) provided written informed consent. Participants were informed that recordings would be released publicly and were encouraged to get assent from others, such as family members or friends, who might appear in the recordings. This project started during the Covid-19 pandemic when outside persons were prohibited on our campuses. Therefore, a sizeable number of recordings were made by the authors of this paper, trainees in our labs, and the persons in our "pandemic bubbles". Trainees were compensated their normal hourly rate or salary for data collection. Outside observers were compensated at rates determined by the common standard of each institution.

The videos were recorded between October 2020 and August 2023 and ranged from one to 73 minutes in length (mean: 19 minutes). Each session is composed of three primary sensor streams: (1) first-person egocentric video from a head-mounted camera, (2) videos of the left and right eye for use in gaze tracking, and (3) information from a tracking camera, including accelerometry, odometry, and gyroscope for use in head tracking. In general, VEDB sessions include all three of these data streams, each of which may be of interest to researchers individually or may be combined to examine the interaction of visual experience with gaze and head/body movement. In addition, we provide processed gaze information for most sessions and an annotation file detailing the tasks and environments depicted in each session.

There is considerable variability in the activities that were recorded. 351 sessions were recorded indoors, and 278 were recorded in outdoor locations. 407 sessions were deemed "active," with observers walking, jogging, skateboarding, or playing other sports, and 222 sessions depicted sedentary activities. Twelve of the 16 top-level categories from the American Time Use Survey (ATUS) were represented. These include personal care, household activities, caring for others, work, education, consumer activities, professional services, eating and drinking, leisure, sports, volunteer work, and travel. Figure 1b shows a word cloud of all recorded activities. The locations where our participants recorded also varied considerably. Of the 365 scene categories in the Places database (Zhou et al., 2017), 124 are represented in the Visual Experience Dataset, see Figure 1a. As these tasks and activities represent self-chosen records from our participants, there is nothing that, if viewed directly, is likely to be considered offensive, insulting, or threatening.

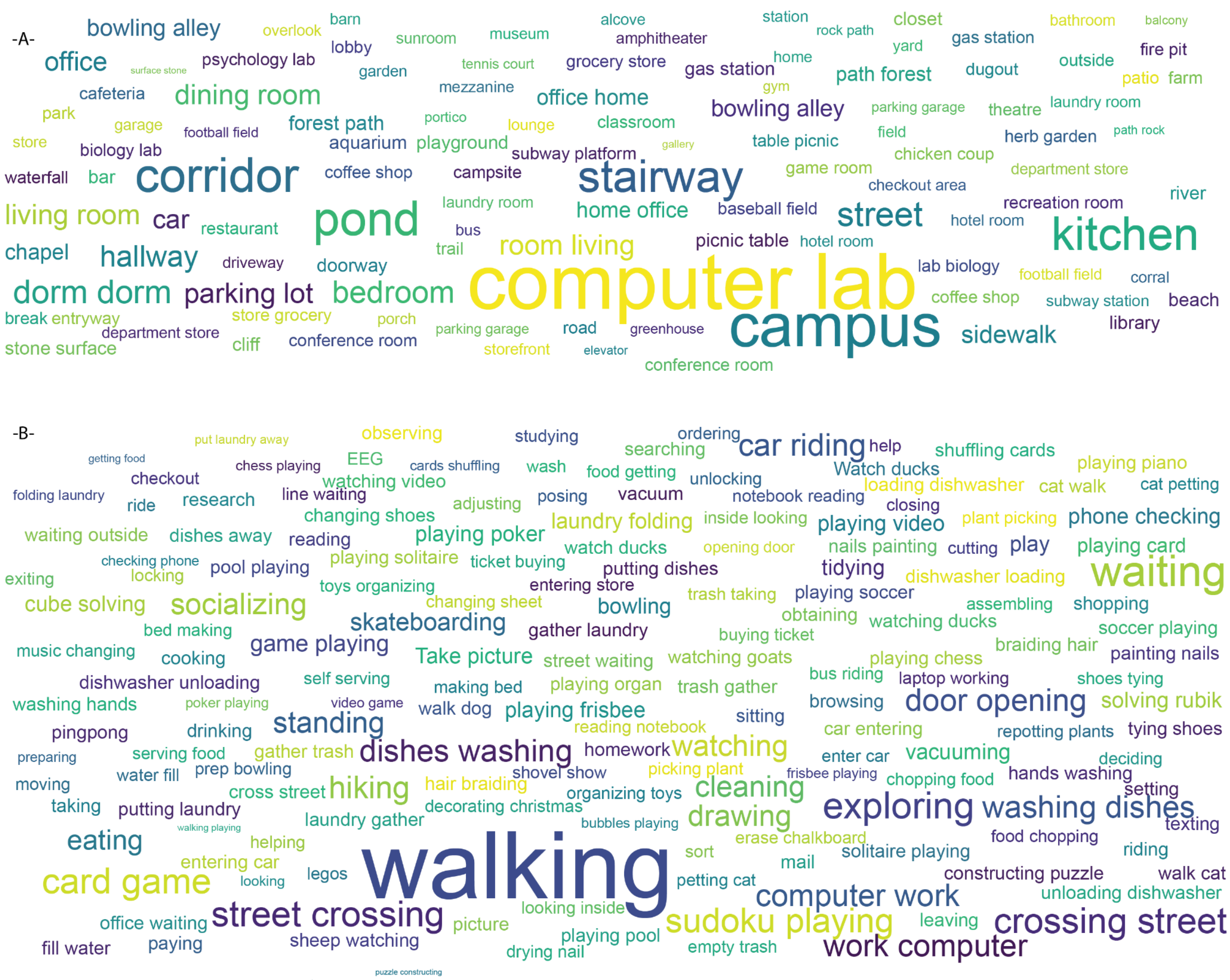

**Figure 1: (A) Word cloud of recorded locations. (B) Word cloud of verbs from recorded tasks. In each, word size is roughly proportional to the number of recorded hours.**

# Collection Process

## Hardware

**Eye tracking**

We used an eye tracking module provided by the *Pupil-Labs* Core system, an open-source and portable device used in various research and commercial settings[1]. The Core system includes binocular eye cameras with 400x400 pixel resolution and a frame rate of 120 frames per second. Eye camera position can be adjusted per participant via a sliding extension arm attached to the device frame, and the Core system headset is lightweight (approximately 23 grams).

**World camera**

The *Pupil-Labs* Core system is available with a small world camera attached to the device frame. However, we attached a different world camera to the Core frame to support higher-resolution first-person video recording with a wider field of view and a global shutter to enable V-SLAM algorithms. A higher resolution video with a wider field of view contributes to richer recordings of first-person experience, including more of the visual periphery and the potential to measure high spatial frequency features in natural scenes. By using a camera that includes a global shutter, we can support more robust measurement of spatiotemporal features in first-person video, such as optic flow or linear filters tuned to specific passbands in the spatial and temporal domain. By comparison, a rolling shutter captures appearance over time by scanning the image row by row, leading to distortions in the appearance of objects that move or otherwise change over time, especially within the scan rate of the device. We incorporated a FLIR Chameleon 3 (Wilsonville OR, USA) camera into the eye tracking device to meet these goals[2]. This camera increases the headset's weight by 30-40 grams but addresses our spatial and temporal video resolution goals. The world camera recorded at a rate of 30 Hz at an image resolution of 3.1 megapixels (2048 x 1536 pixels). Some sessions (N=166) employed a 3.6 mm 1 /2.5” lens to this camera. This enabled a horizontal field of view of 125° but had a very heavy fisheye distortion and lens vignetting. Other sessions (N=551) employed a 4 mm 1/1.8” lens, with a horizontal extent of 101°. This provided a more natural-looking video while still recording a larger field of view than is typical (see Table 1). Figure 2 shows the relative vignetting of both lenses. We recorded five seconds of video of a uniform white wall, converted these frames to grayscale, and examined the relative pixel intensities across the visual field. Given the heavy vignetting of the 3.6 mm lens, we do not recommend measuring natural scene statistics in the periphery for these sessions. That said, the 3.1-megapixel resolution provides a sufficient number of pixels to compute image statistics from center-cropped portions.

---

[1] Full specifications: https://pupil-labs.com/products/core/tech-specs
[2] Full specifications: https://www.flir.com/products/chameleon3-usb3/?vertical=machine+vision&segment=iis

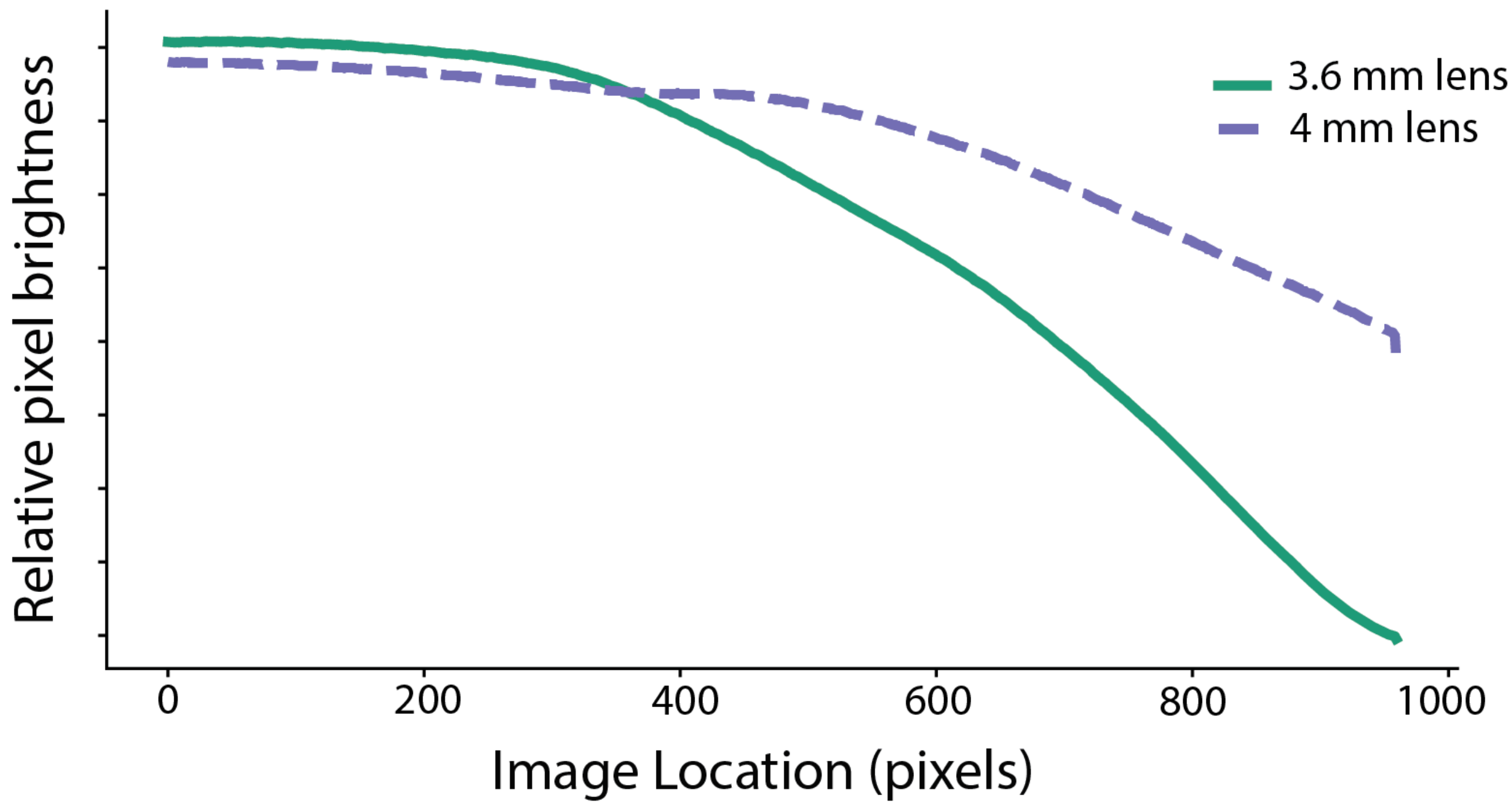

**Figure 2: Lens vignetting for 3.6 and 4 mm lenses. Five seconds of video of a uniform, white wall was recorded with each lens. The video frames were converted to grayscale, and the pixel intensities were plotted as a function of horizontal location. Zero represents the center of the image plane.**

### Tracking camera

We wished to record head movements in addition to egocentric video and eye movements to allow research questions linking natural scene statistics to head movements associated with walking, running, or other aspects of body movement related to visual exploration or the completion of everyday tasks. We used the Intel RealSense T265 as our head tracking module to record odometry information[3]. This device has stereo wide-angle monochrome cameras, and an inertial measurement unit (IMU) consisting of a gyroscope, and an accelerometer. Data from these sensors can be used to infer the device (head) pose in the world via a proprietary visual-inertial simultaneous localization and mapping (VI-SLAM) algorithm. The T265 estimates linear and angular head position via VI-SLAM at 200 Hz, which can be differentiated into velocity and acceleration. Furthermore, the T265 provides supplemental estimates of angular velocity and linear acceleration from the triaxial gyroscope and accelerometer on the device, with sampling rates of approximately 200 and 60 Hz, respectively. The T265 added 55 grams to the weight of the recording device.

[3] Full specifications: https://dev.intelrealsense.com/docs/tracking-camera-t265-datasheet

**Headset designs**

We needed to design custom headsets to include each of these sensor components, mount them rigidly with respect to one another, and fit on observers' heads stably and comfortably, allowing children and adults to wear them for an extended period while performing everyday activities. Through iterative design and redesign, we found that no single solution worked best for all participants in all activities. Adult participants varied substantially in head size and face shape. Further, children required smaller, simpler headset designs to accommodate their size and patience. Thus, the final dataset reflects data collected from four different headset designs. We will briefly describe each, highlighting each design's possible advantages and disadvantages.

*Hard plastic mount*

We used a ratchet-adjustable hard-hat suspension (3M, St. Paul, MN. USA) as the base of this headset (See Figure 3 (left panel)). We designed custom 3D-printed pieces to hold the world camera (FLIR), the tracking camera (Intel), and the eye tracker (Pupil) rigidly to one another. We used multiple 3D printed clips to ensure that the weight of the world and tracking cameras were evenly distributed over the device and secure enough to remain stable when a participant was in motion. While this design achieved the goal of rigidly positioning the cameras relative to one another and to the participant's head, the design did not flexibly accommodate all head sizes and face shapes. Multiple observers found this design uncomfortable because the device had to be angled upward to keep the observer's eyes in the Pupil camera frame. To keep the device stable in this orientation, the headband had to be very securely tightened, reducing the comfort of the headset.

*Elastic strap mount*

We used elastic headbands designed for GoPro cameras (see Figure 3 (center)) for this design. As before, we used custom 3D-printed parts to mount the cameras relative to rigidly to one another and to the elastic mount system. This system allowed users to adjust the circumference and longitude of the mount and was thus more flexible and more comfortable. However, this system was difficult to adjust, and the elastic mounting may have allowed the system to slip or bounce during vigorous movement. Further, the silicone coating of the elastic straps pulled the hair of some observers. We created a variant of this mounting system to mitigate this discomfort by pulling the elastic straps over a standard bicycle helmet. We used duct tape to affix the Pupil Core frame to the helmet. While this variant was comfortable to wear, the increased camera height reduced the visibility of the lower visual field.

*Virtual reality headset mount*

We used a Quest 2 Elite mount (Meta, https://www.meta.com/quest/accessories/quest-2-elite-strap/ ) combined with custom 3D-printed parts (see Figure 3 (right)). This mounting system included a ratchet-adjustable plastic strap that was wider than the previous plastic mount, increasing the comfort for some observers. Further, unlike the previous plastic system, the Quest 2 mount included an elastic strap running from over the top of the head front to back, increasing the system's comfort by distributing the weight of the front-facing cameras more evenly. Although some observers found this model comfortable and stable, there were fewer degrees of freedom to adjust the system, and not all face sizes and shapes were easily accommodated.

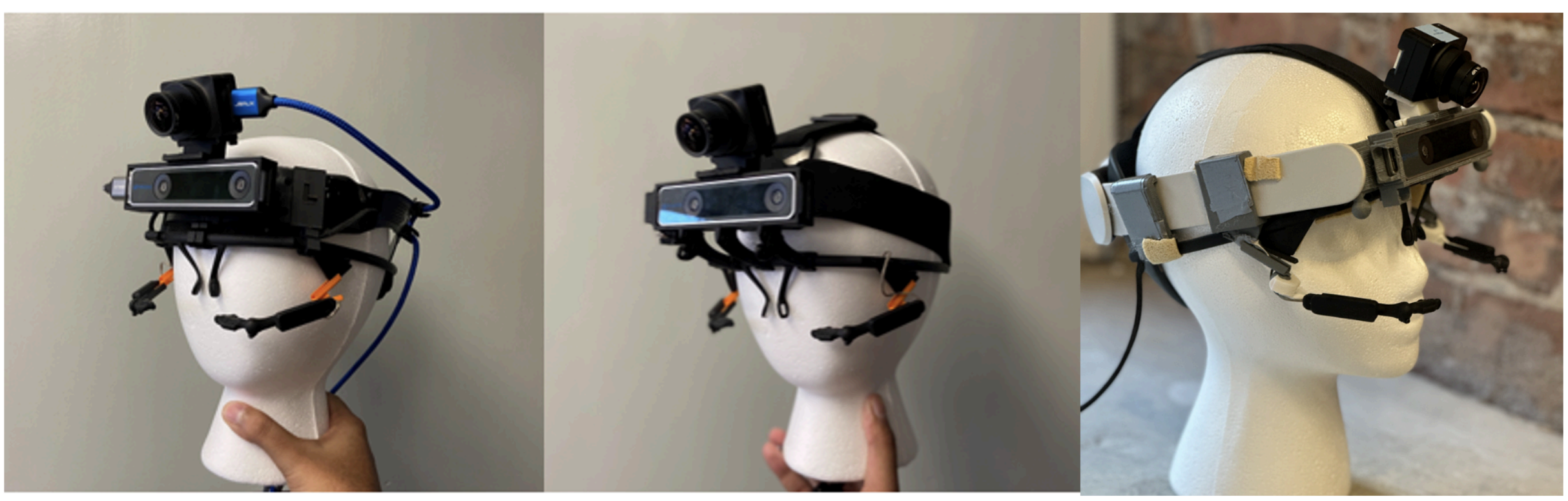

**Figure 3: (Left) hard plastic mount; (Middle) elastic strap mount; (right) Virtual reality headset mount. The right mount also shows the custom ball joint eye extenders.**

*Custom Pupil Camera Extenders*

The Pupil Core system is equipped with sliding extenders that allow the eye camera to adjust to different face shapes. However, these extenders are at a fixed angle, and we found that this angle only accommodated some observers. With this system, many observers were forced to place the entire device at an unusual angle for their eyes to stay in the camera frame, which could be uncomfortable or unstable. We designed and printed a new extender that employed a ball joint mechanism, enabling each user to customize their device, and leading to higher quality data. This joint was susceptible to wear and loosened with use, requiring frequent replacement.

*Peripherals*

The eye tracking, world, and tracking cameras were each plugged into a laptop computer via USB, and the computer was placed in a backpack. Given the intensive battery demands of recording four video streams simultaneously, we employed a backup computer battery for longer sessions. Participants were provided an Amerteer Bluetooth™ clicker module to pause and resume recording if needed. Calibration and validation targets were provided on a printed display (see below) or a smartphone.

**Acquisition software**

We developed custom code to record the four video streams as well as odometry and IMU data from the tracking camera. We recorded timestamps for each stream on a common clock to account for the different frame rates of each camera. The acquisition code was based on PupilLabs' open source code for their PupilCapture software (https://github.com/vedb/pupil_recording_interface) , modified to incorporate the FLIR and Intel T-265 cameras. To economize file sizes, all video data was compressed at the point of saving to disk using the h264 codec, with a constant rate factor of 18. This parameter value is commonly considered to avoid introducing perceptible compression artifacts (e.g. (Wang et al., 2017)), while reducing the amount of data saved to disk for each session from approximately 24 GB / min to  approximately 0.4 GB / min, a 60x reduction in file size. (Note that the benefits of compression will vary with the frame-to-frame differences in different videos.)

The variable illumination of VEDB recording locations presented a substantial technical challenge. To account for these difference, we allowed the frame rate of our world and eye camera acquisition to vary over individual recording sessions, collecting light for slightly longer in dimmer conditions. Frame time was also slowed for stretches of some sessions due to low battery levels or processing demands in the recording laptop. We documented this variability by collecting frame acquisition timestamps for all video streams. These are saved separately for each session and stream as a Numpy file. These files are critical for basic statistics such as motional statistics in the world camera, or defining saccades based on a velocity threshold. The code for the acquisition software is available on GitHub (https://github.com/vedb/pupil_recording_interface and https://github.com/vedb/ved-capture).

## Acquisition Procedure

Each participant provided written informed consent (or, in the case of minors, assent with parental written consent). To ensure safety, we instructed participants not to drive and to remain aware of their surroundings. To protect the rights of bystanders, we instructed observers to obtain permission from acquaintances who were filmed in private spaces and to stop recording immediately if requested by someone in a public place. We informed observers that recordings would be made available on a public-facing website and were allowed to delete any session or section for any reason.

We trained observers to properly wear and adjust the recording equipment, use the custom recording software, and complete the calibration and validation procedures. This training typically took between 20 minutes and two hours, and observers were compensated for their training time.

The eye tracking calibration target consisted of a standard concentric circle pattern provided by Pupil Labs. This target was either printed on paper and affixed to a 3D-printed wand (7.5 cm by 13 cm in size) or viewed on a 6.5 cm by 13 cm sized smartphone. The validation target was a checkerboard pattern printed and affixed to the opposite side of the calibration wand or viewed on a smartphone. Some recordings used a 7x9 checkerboard pattern, while others used 4x7.

We instructed observers to hold the calibration target at arm's length and to reveal the target at 25 locations, evenly spaced in a 5x5 grid in the world camera's field of view. We directed observers to hold the calibration target steady, keeping a steady gaze at the center of the centric circle pattern for 1-2 seconds in each location. As observers had no real-time feedback of their calibration target locations, some targets were missed because they were out of the world camera's field of view. To avoid spurious detections of the calibration target, observers were advised to rotate the calibration target by 90° (to view the calibration wand edge-on) while moving the target between locations. The validation procedure was identical, but used the checkerboard target instead of the concentric circle target. To calibrate the tracking camera, observers nodded slowly five times and then rotated their heads right and left five times while regarding the validation target. These movements may be used to identify a head-centered kinematic reference frame that is common across observers and independent of how the device is mounted on the head; they also provide a basis for joint analysis of head and eye movements. Not all sessions contained tracking camera calibration. Some sessions also contained additional validation procedures at other stages of recording to evaluate the stability of the initial eye and head tracking calibration. Following the session, observers were invited to view their recordings before uploading them to a centralized server. In the debriefing form that participants received after the experiment, they were provided with contact information for the PI should an observer wish to withdraw their data at any time.

## Potential Sources of Error or Bias

**Omission**

In total, 968 sessions (283 hours of data) were recorded from 70 unique individuals, with three universities participating in data collection. Excluding sessions that had corrupted recordings (N=59), contained only pilot data (N=76), or were disallowed by institutional IRB (N=116), we have 717 sessions (244 hours) from 56 unique individuals.

**Video Recording Errors**

Sixteen sessions have been included but with minor recording errors. These include a blurry world camera (N=3), partial occlusion by a cord or other part of the recording apparatus (N=6), and an inability to maintain a sampling rate near 30 Hz on the world camera (N=5).

**Eye Tracking Errors**

We could successfully calibrate gaze tracking in 548 of the 717 sessions (76%). Of the 169 sessions without calibration, 16% were due to the participant omitting the calibration procedure. Unfortunately, gaze will not be recoverable from these sessions. Another six sessions had corrupted timestamp files for one or more of the cameras, which are also non-recoverable errors. The remainder of the errors are due to difficult lighting or backgrounds during calibration and may be recoverable in future by improved marker detection algorithms or manually annotating the calibration target locations. We are presenting gaze error as the error in the independent validation sessions. To compute a representative error metric, we employed a series of procedures to omit marker detections that were likely to be spurious. We removed marker detections that were very brief. These were often due to false alarms with the scene background, or views depicting a target that was not intended to be part of a validation sequence (such as a target left upon a table). In addition, we spatially clustered validation marker locations, and only computed error for sessions with a sufficient number of validation points. However, the clustering procedure identified zero clusters for some sessions. Therefore, the final validated gaze error values are from 458 sessions (64% of total, 84% of sessions with successful calibration). We annotated the reasons for each failure. For 110 sessions (65% of failures), insufficient information was recorded to recover gaze. These include omitting calibration or validation (N=52), corrupted timestamp files for any of the video streams (N=8), pupils that were not entirely visible to the camera (N=22), or validation that was not correctly executed (N=28). The remaining errors were due to challenging backgrounds and lighting conditions that made it difficult to detect the validation target automatically. Figure 4 shows a diagram of the validation failures and their reasons.

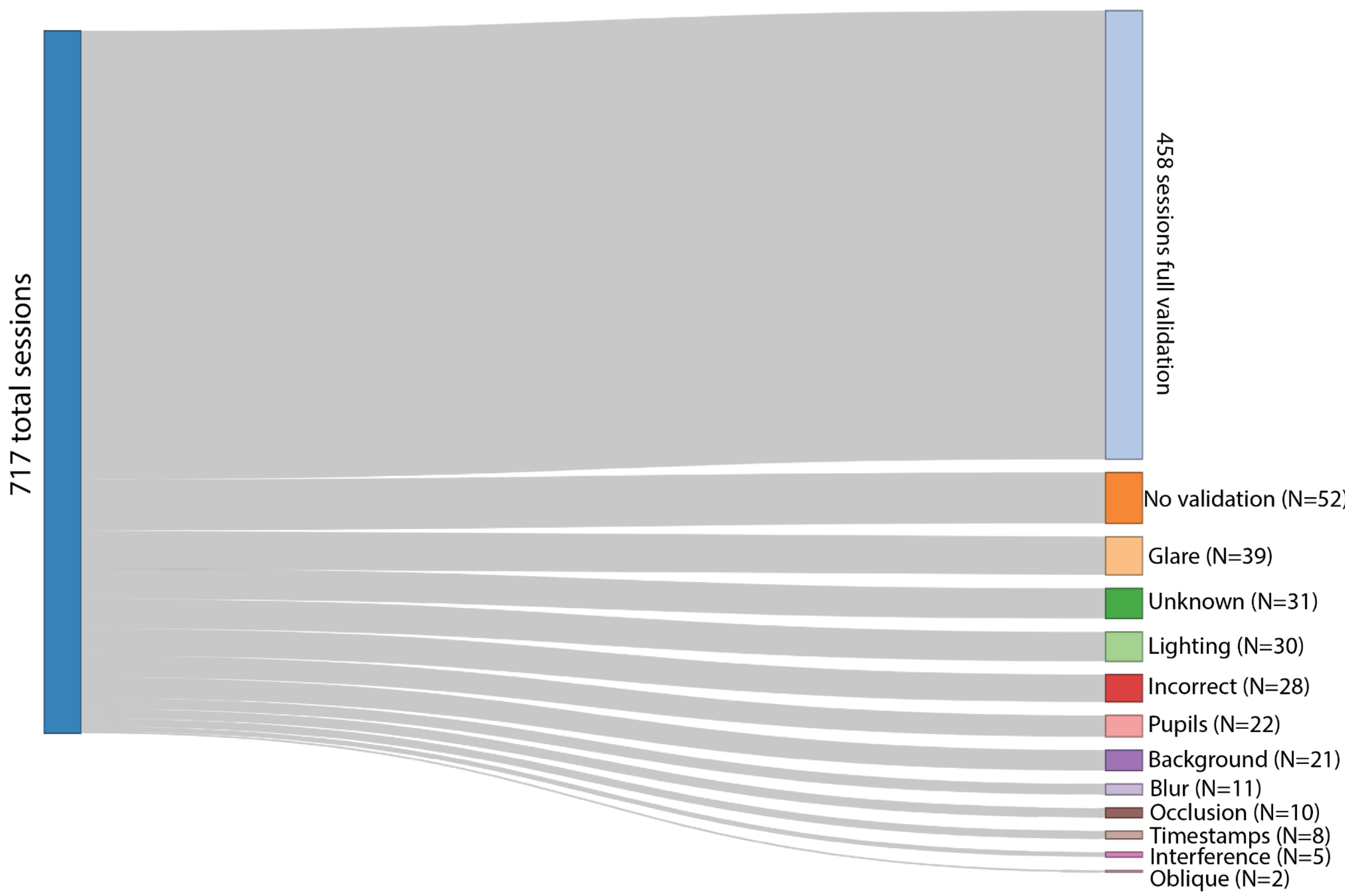

**Figure 4: Sankey diagram indicating reasons for validation failures. About one third of these failures may be recoverable with additional annotation.**

The extent to which the pupil can be confidently identified limits gaze tracking accuracy. We measured the proportion of frames with PupilLabs confidence ratings above 0.6 in each session. We found that, on average, 65% of frames met this confidence threshold. However, the type of recording significantly impacted confidence ratings, as shown in Figure 5. Indoor sessions led to more retained frames compared to outdoor sessions (68% versus 62%, Welch's two-sample t-test (t(593.7)=4.78, p<<0.001)), and sedentary sessions led to more confident pupil detections than active (70% versus 63%, t(347.4)=-4.03, p<<0.001).

-A- -B-

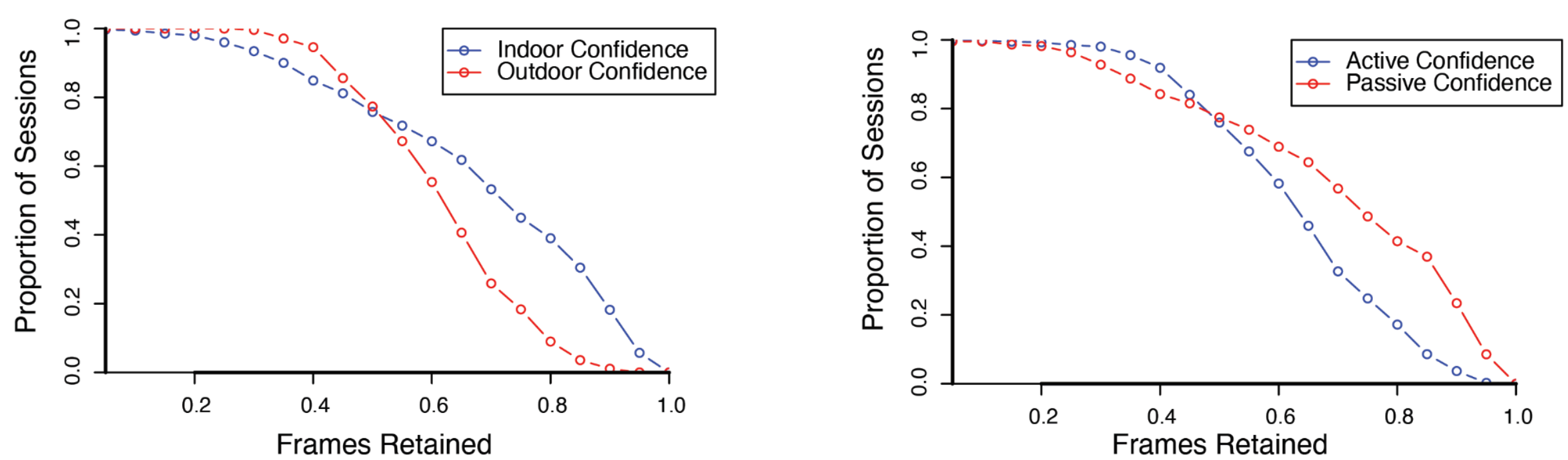

**Figure 5: (A) Survival plot of sessions, plotting the proportion of sessions on the Y axis as a function of frames retained (frames with pupil detection confidence > 0.6). Indoor sessions led to significantly higher pupil confidence than outdoor sessions. (B) Survival plot for active and sedentary sessions. Sedentary activities led to significantly higher pupil confidence.**

The validity of the initial calibration requires that the cameras remain in the same location relative to one another and in the same position on the observer's head. While the mounting systems guarded against the first issue, any device instability on an observer's head could lead to inaccurate gaze tracking. To assess the extent to which stability limited gaze accuracy over time, we used OLS regression to fit a model of each pupil's y-position over time and saved the slope parameter. Figures 6A and 6B show histograms of slopes for the left and right eyes. These distributions are centered on zero, with very small slope values. However, small differences in slope could lead to large errors in gaze tracking, particularly in long sessions. Thus, to quantify the absolute magnitude of drift from beginning to end of each session, we averaged the two eyes and converted the difference in average position between the beginning and end of each session to degrees of visual angle. As shown in Figure 6C, 95% of sessions contained less than five degrees of drift (70% of sessions drifted by less than two degrees). Active sessions had a numerical tendency to greater drift compared to passive (1.83 versus 1.51 degrees, t(396.1)=1.93, p=0.054). We observed no significant difference in drift between indoor and outdoor sessions (t(510.7)<1). Finally, we reasoned that if slippage were a serious issue in many sessions, then it should get progressively worse with longer sessions. To test this, we computed a correlation between session length and amount of drift. We observed a small *negative* correlation between session length and the amount of drift (r=-0.09, p<0.05), indicating that slippage did *not* get worse with longer sessions. It is possible that drift followed nonlinear patterns, with sudden shifts in headset position over short times, which our linear regression estimation did not fully capture. Estimation and correction of this level of slippage is beyond the scope of the current manuscript, but the linear regression and drift magnitude analyses presented here, taken together, suggest that for a clear majority of sessions, gaze calibration remained stable over time.

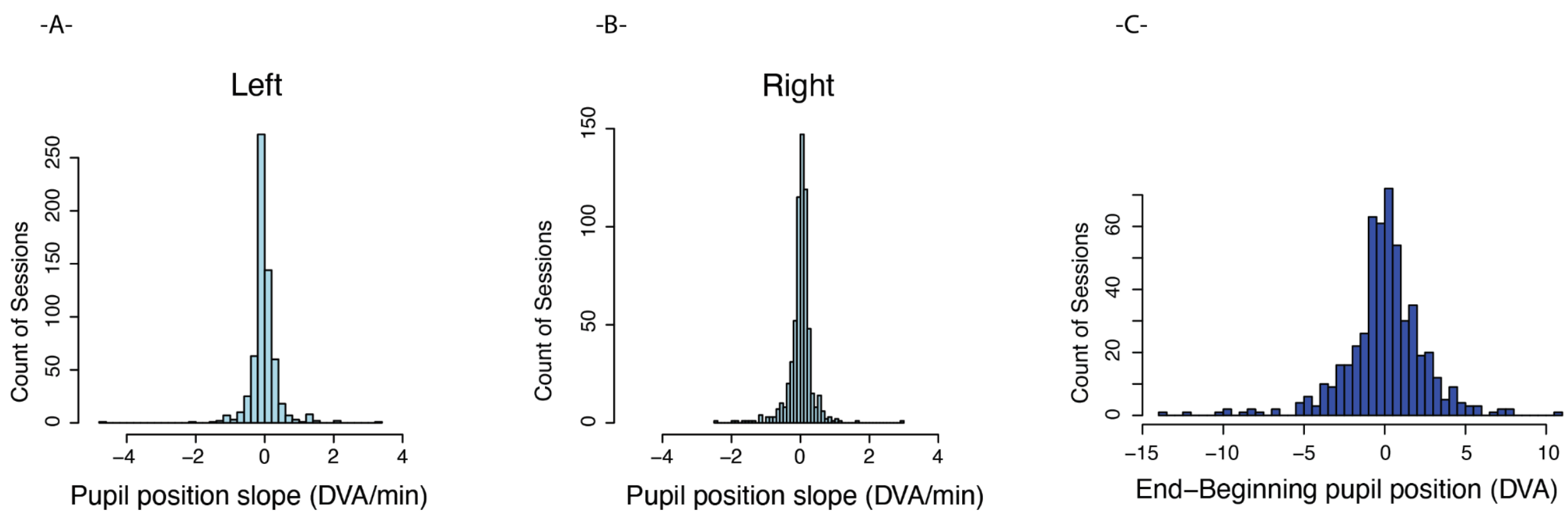

**Figure 6: (A) and (B) slope of detected pupil positions over time in units of degrees of visual angle (DVA) per minute. (C) Difference in average pupil position between early (first 1000 frames) and late (last 1000 frames) in each session, expressed in degrees of visual angle.**

Finally, we assessed the consistency of temporal sampling in each stream by computing the mean squared error (MSE) of sample timestamps relative to an assumed constant frame rate. MSE values were generally very low (specific values can be found in Table 2), indicating that our sampling was stable over time.

| **World** | **Eye0 (right)** | **Eye1 (left)** | **Accelerometer** | **Gyroscope** | **Odometry** |
|---|---|---|---|---|---|
| 9e-5 | 1.8e-6 | 1.9e-6 | 0.0004 | 4.3e-5 | 4.7e-5 |

**Table 2: Mean squared error (MSE) values interframe intervals of each data stream.**

## Confidentiality and privacy

Our lived experiences contain many potentially sensitive moments. We have mitigated this risk to the extent possible by allowing each observer to choose what activities to record. Further, the use of the Bluetooth pause button allowed participants to pause recordings during sensitive times, such as entering a password or passing through a room with a family member who did not wish to be recorded. Finally, observers could note sessions or session sections that they wished to be removed for any reason. Nonetheless, it is possible that some potentially sensitive information may exist in the dataset.

We have taken measures to reduce the identifiability of individuals within the dataset. Even simple demographic information such as zip code, gender, and date of birth can be used to identify individuals uniquely (Sweeney, 2000). We have reduced these risks by recording each participant's date of birth as January 1 of their birth year and not recording the location of any specific recording. However, we recognize that the unique landscapes of our recording locations make it possible to make educated guesses about the location.

Finally, the eye videos themselves may be considered sensitive. Near-infrared videos, such as those used by the Pupil Core, have been shown to contain diagnostic iris features that can be used for biometric identification (John et al., 2019). As the eye videos contain the raw data for researchers advancing new pupil detection and gaze tracking methods, the privacy interests of the participants are opposed to the scientific utility of these videos. One study has shown that a mild degree of Gaussian blurring of these videos can remove diagnostic iris information without degrading pupil tracking performance (John et al., 2019). Therefore, we have released blurred versions of the eye videos to improve the impact of the dataset without harming participant privacy. Unmodified eye videos are made available to researchers with a Databrary account.

## Sampling Strategy

Our sampling strategy considered several tradeoffs. First, we wanted data to be representative of daily experience but also to capture the breadth of activities that are common across many groups. According to the American Time Use Survey (ATUS), an annual telephone survey conducted by the United States Bureau of Labor Statistics, the preponderance of peoples' days are engaged in work (7.69 hours, on average in 2022) and sedentary entertainment such as watching television or playing computer games (over 3 hours, on average). As some occupations cannot be recorded for privacy or safety reasons, we have limited the "work" task to common office computer work in the dataset. Television and video games are common but visually homogenous activities, so we avoided sampling them in proportion to the time most people allot to these experiences.

Second, we wanted to show a variety of people engaging in various tasks in various environments without breaking the natural covariation between people, tasks, and environments. To accomplish this aim, we allowed participants to determine what they wanted to record in consultation with experimenters to assess the safety and feasibility of each activity.

Third, we wanted to record from various observers and tasks while also enabling researchers to study interesting repeated measures, such as similarities within individuals or across individuals performing the same task in the same locations. To accomplish this, we included some structured instructions within our sampling. For example, 12 observers were instructed to walk around a 0.25-mile pond path on the Bates College campus once or twice a week for the duration of 2022. These 135 sessions can be useful in assessing eye and head movement variability within and across individual observers. Figure 7 shows example frames of representative activities in the VEDB.

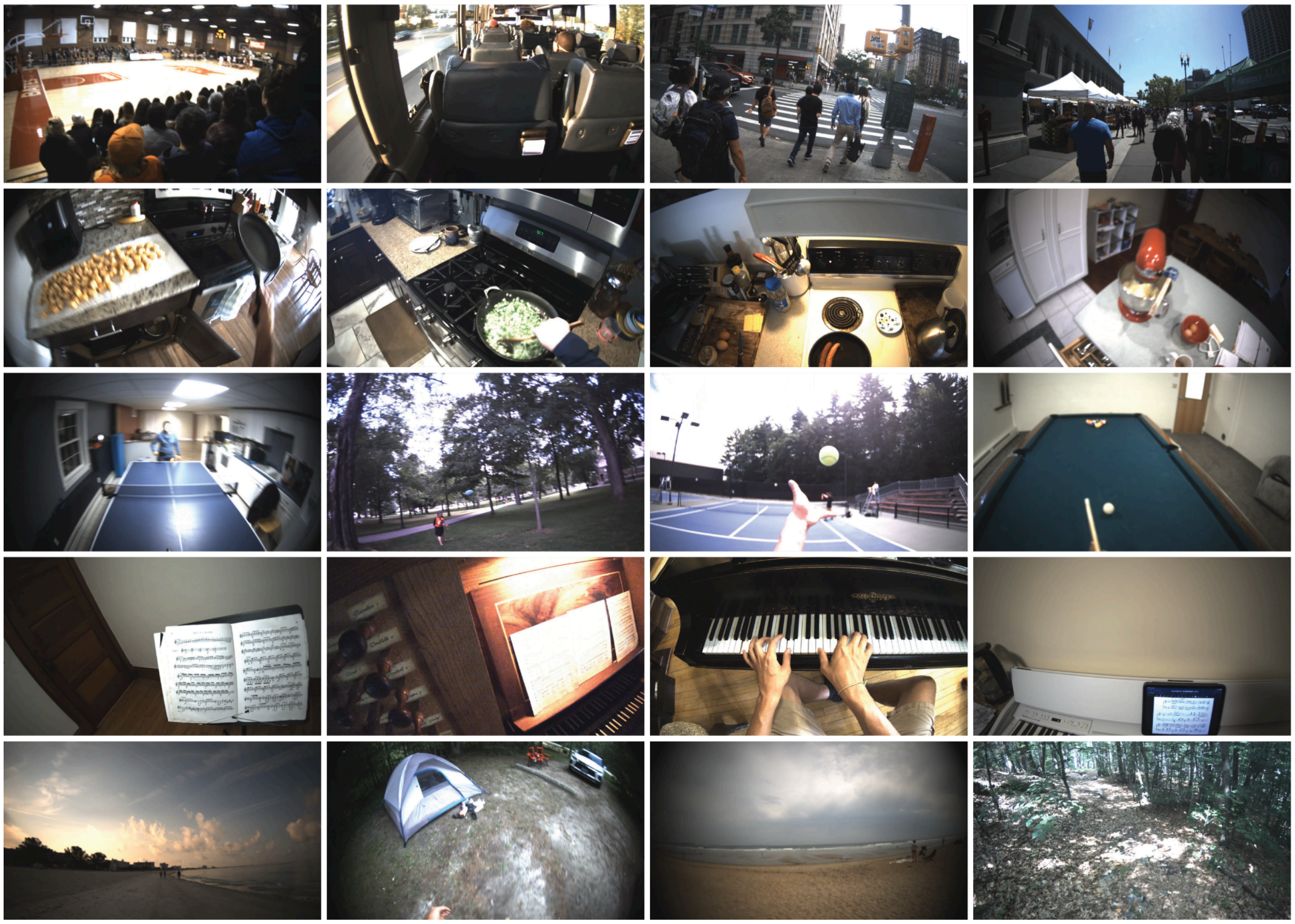

**Figure 7: Representative frames within the VEDB. Top row: public activities; Second row: cooking; Third row: playing music; Bottom row: outdoor activities.**

# Data processing

### Preprocessing

Using custom software, we hand-annotated the start and end times for eye movement calibration, validation, and odometry calibration procedures. These times were translated into timestamp indices and saved in a .yaml file (marker_times.yaml).

### Gaze pipeline

*Calibration Target Detection:* We applied the circle detection algorithm from Pupil Labs to detect the concentric circle pattern in the world videos between the saved calibration start and end times. The locations of each detected circle, along with their size (for filtering spurious detections), and time stamps were saved as NumPy arrays.

*Pupil Detection:* We applied the 2D pupil detection method from Pupil Labs to locate the pupil in both eye videos. We saved the detected pupil ellipses and their diameters, locations, luminance, and time stamps in Python dictionaries. Further, the confidence of each detection (0-1) was saved as a NumPy array.

*Calibration:* The concentric circle calibration target locations were filtered before the calibration step for spurious detections. Detected locations that were too small, at an inappropriate aspect ratio, or detected very briefly (< 0.3 s) were omitted as likely to be spurious. Target locations were not completely steady because calibration targets were held by hand. We clustered the remaining locations using DBSCAN clustering to increase the signal-to-noise ratio. This method was chosen as the number of clusters does not need to be specified in advance, and it is robust to outliers. Pupil locations were filtered by confidence, with only confidence values above 0.6 included in the calibration. To account for unsteady hands and occasional behavioral gaze errors (e.g., participants momentarily looking away from the target due to wind or other real-world distraction), we used the median location within each cluster for both calibration marker detections and detected pupils to compute gaze calibration. Next, the saved calibration target arrays and pupil arrays were passed to custom Python code that fit the pupil locations to each calibration location using cross-validated thin plate spline regression. Model fitting was performed monocularly.

*Computing validation error:* We detected the locations of the validation markers using the checkerboard detector from the OpenCV library and stored these as NumPy arrays. As with the calibration markers, we filtered these detected locations by size and duration and clustered the locations using DBSCAN. Validation target locations over four standard deviations away from cluster medians were omitted as outliers. The error (in degrees of visual angle (DVA)) between the modeled gaze and the location of the checkerboard's center was computed and stored. A histogram of validation error is shown in Figure 7b. 168 sessions have a validated gaze error of less than two degrees, and 332 sessions have validated error under five degrees. Indoor and outdoor recordings had similar levels of validated error (indoor: 2.4 DVA, outdoor: 2.6 DVA, $p=0.15$ Mann-Whitney U). Similarly, we observed no significant difference between physically active versus passive sessions (active: 2.6 DVA, passive: 2.4 DVA, $p=0.48$).

*Computing immediate calibration error:* The checkerboard pattern was not robustly detected in some environments and lighting conditions. We also computed the error in gaze calibration to estimate the extent of gaze error for these sessions. This quantity is overfitted to the data but can be used to identify sessions with acceptable gaze calibration. Using the detected calibration markers, we used affinity propagation to cluster marker locations and then computed the median location of each cluster. We identified the time stamps associated with each cluster and computed the median gaze position during that same time interval. We computed the degree of error between them (measured in degrees of visual angle (DVA)). As shown in Figure 8a, the median calibration error in most sessions was minimal (59% were <1° and 80% were <5°). There was a long tail of very high error clusters. There are numerous reasons for these failures, including spurious pupil detections, pupil occlusion through eyelashes, or an eye that was not centered in the frame. We observed a slight tendency for outdoor sessions to have a higher median calibration error compared to indoor (0.65 versus 0.40 degrees of visual angle, Mann-Whitney U p<0.05). This was driven by the additional challenges of recording outdoors, including participants squinting into the sun. We observed lower median calibration error for physically passive sessions compared to active (0.32 versus 0.63 DVA, p=0.006). The code that performs this gaze pipeline can be found on GitHub (https://github.com/vedb/vedb_gaze_1.0).

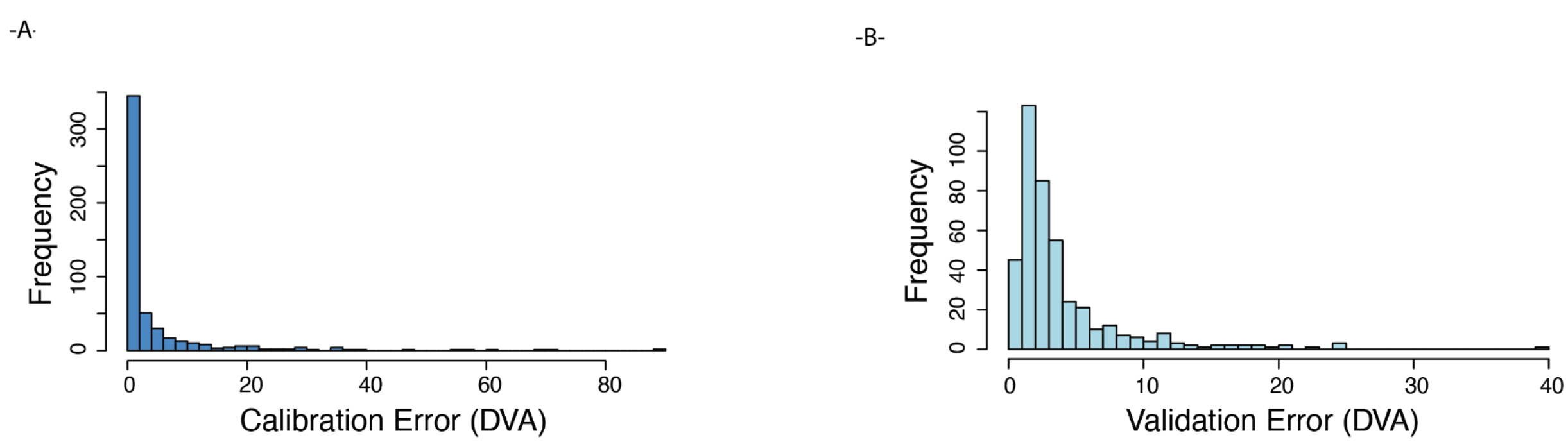

**Figure 8: (A) Histogram of initial calibration error (degrees of visual angle, DVA); (B) Histogram of validation error (DVA) across 423 sessions.**

### Head-tracking

Odometry, gyroscope and accelerometer data are provided as is without further processing. These raw data indicate the position and movement of the tracking camera relative to a world reference frame that is initialized when recording starts. Precision and accuracy of this tracking data relative to in-lab optical tracking has been assessed previously (Hausamann et al., 2021). For the purposes of comparing or compiling head movement data across participants, it is necessary to transform the data into an anatomical or kinematic reference frame that is the same across participants and independent of how the device is mounted on the head (Sinnott et al., 2023). Code for finding and applying this transformation for a given session can be found on GitHub (https://github.com/bszek213/odopy). Similarly, for joint analysis of head and eye movements, or for reconstructing how the eye is moving relative to the world, the spatial relationship between the reference frame of the tracking camera and the world camera is necessary because eye movements are measured in the reference frame of the world camera (Hausamann et al., 2020). Code for finding and applying the eye-head reference frame transformation for a given session can be found on GitHub (https://github.com/bszek213/head_eye).

**Labeling**

For each session, we extracted frames from the world camera videos at five-second intervals. Researchers in our groups examined these frames and, for each session, created a .csv file that labeled all changes in scene location (using scene categories from the Places database) or the observer's task in the session. This provides not only ground truth place and task labels but also the approximate time points of each scene or task transition.

**Preserving second-order privacy**

We created custom software to blur bystanders' faces in our videos. Each video frame was read in and downsampled to 50% size for speed. We applied the RetinaFace detector (Deng et al., 2020) to each frame. We created a tight ellipse around each of any bounding boxes returned by the detector and blurred the pixels within the ellipse with a Gaussian blur (SD=149 pixels). We then rewrote each altered frame to a new file. We found that RetinaFace provided the best detection efficacy of several leading detectors. Specifically, while most detectors underperformed on non-white and non-adult faces, we found that RetinaFace was robust across age, race, and the presence of face masks. An internal audit of about 3000 of the most challenging sessions found that the classifier's sensitivity was over 87%, making it robust across different ethnicities, ages, viewpoints, and occluders such as hats, sunglasses, and face masks. A sample of 150 typical face frames yielded a sensitivity of over 99%. The script that generated these videos is found on GitHub (https://github.com/vedb/protect_privacy).

## Use cases for the VEDB

The VEDB is appropriate for studies in natural scene statistics, examinations of gaze behavior during common tasks, and studies of how head and eye movements combine to orient overt attention and gaze. The VEDB also provides a large dataset of high-resolution egocentric video together with head and eye movement data. Alongside investigations of natural scene statistics, these data can be used to create and validate new or extant methods of motion or position estimation, object classification, human activity recognition, machine learning-based eye tracking, and more. Below, we provide our plans for this dataset and our recommendations for others using the dataset.

One unique feature of the VEDB is that scene locations and observers' tasks have been annotated at a fine temporal scale. We have been working to assess the relative success of pretrained deep neural networks (dCNNs) in classifying scene examples in the VEDB. We have found that these networks are impaired in classifying VEDB samples and that this effect is particularly strong for scenes within private homes (Greene et al., 2022). Thus, VEDB frames may be useful for fine-tuning dCNNs to reduce their biases. Further, the fine-grained annotations of tasks can be useful in training systems to recognize human activities from an egocentric perspective, which can be useful in human-computer interaction and robotics.

The majority of the literature on natural scene statistics has only considered static photographs. Thus, the VEDB is especially appropriate for studying spatiotemporal image statistics. Sub-sampling of world camera images centered on gaze allows generating retino-centric videos that provide an approximation of moment-to-moment retinal image motion. Furthermore, by combining the world camera feed with the odometry from the tracking camera, one can begin to assess the statistics of self-generated and externally generated visual motion. The statistics of head orientation and movement have important implications for understanding sensory processing across vestibular, visual and auditory modalities (MacNeilage, 2020; Sinnott et al., 2023), and joint statistics of natural head and eye movements can provide valuable and novel insight into head-eye coordination during everyday behavior. Because each session is named with a recording time stamp (YYYY_MM_DD_HH_MM_SS), we have been examining how local environmental conditions (temperature and time until sunset) alter some standard natural scene statistics (Greene et al., 2023).

Gaze information is available for most VEDB sessions, with 168 sessions (approximately 63 hours of content) containing initial gaze accuracies of under 2 degrees of visual angle, making them appropriate for many eye tracking applications. For research questions that do not require precise gaze tracking, 333 sessions (~122 hours) have median gaze accuracies of under 5 degrees of visual angle. We are providing blurred eye videos in the dataset, which previous work suggests is an effective method for protecting observers' identities while allowing pupil detection (John et al., 2019). More robust pupil detection methods will improve any future gaze pipeline applied to these data. In particular, machine learning methods show great promise for improving gaze tracking pipelines (Yiu et al., 2019).

The VEDB highlights the challenges faced by researchers who wish to study gaze in active situations with a diverse set of participants. In particular, the VEDB contains gaze data recorded in challenging outdoor conditions with dynamic lighting, which require careful sensor parameter selection with model-based tracking methods (Binaee et al., 2021). Further, the VEDB contains data from participants of many racial and ethnic backgrounds, a known challenge for model-based eye tracking methods (Blignaut & Wium, 2014). Thus, the VEDB provides an opportunity for researchers to develop tracking methods that are robust to these features. Finally, by combining gaze, egocentric video, and positional data, researchers can construct and validate gaze-contingent methods of human activity recognition, a topic of increasing interest in the wearable and augmented reality communities (Bektaş et al., 2023).

## VEDB Case Study

Given the critical role of faces in our social lives, the "face diet" of human observers has been of particular interest. Developmental work using head-mounted cameras has shown that human infants see more faces than hands in the first few months of life, and that this ratio switches as children become toddlers and can engage more interactively with the world (Fausey et al., 2016). Smaller-scale studies of adults have examined the frequency of encountering faces, along with their viewpoints and depth profiles (Oruc et al., 2019). Here, we demonstrate the utility of the VEDB by expanding on this type of analysis by using gaze information in addition to frequency statistics from egocentric video.

We applied the Google MediaPipe (Lugaresi et al., 2019) face and hand detectors to each video frame after calibration and validation. We omitted calibration and validation segments because hands were continuously visible during the procedure. This analysis totaled over 22.5 million frames of content. Heat maps showing the spatial distribution of faces and hands in the world video is shown in Figure 9 (top). Our findings revealed a distinct spatial distribution of these body parts: the median center of faces was located 13 degrees above the video center, while the median hand position was 23 degrees below (Wilcoxon rank sum, $p<0.0001$).

We next applied our gaze data to show the spatial distribution of faces and hands relative to fixation. For each frame with a detected hand or face, we found the gaze location and, when possible, cropped a 512x512 image segment centered on the gaze location. The resulting gaze-centered heat maps are shown in the bottom row of Figure 9. Interestingly, the highest intensity location on the gaze-centered face map was 2.8 degrees above fixation, while the highest intensity location on the hand map was 3.9 degrees below fixation. Taken together, this suggests that while both faces and hands tend to be frequently fixated by human observers, retinotopic experience of these objects still differs spatially. This suggests the hypothesis that face and hand detection sensitivity may vary accordingly across the visual field. We are currently testing this hypothesis. Furthermore, characterizing the retinotopic visual experience of various categories can provide additional insight into the functional organization of higher-level cortex (e.g. (Weiner & Grill-Spector, 2013).

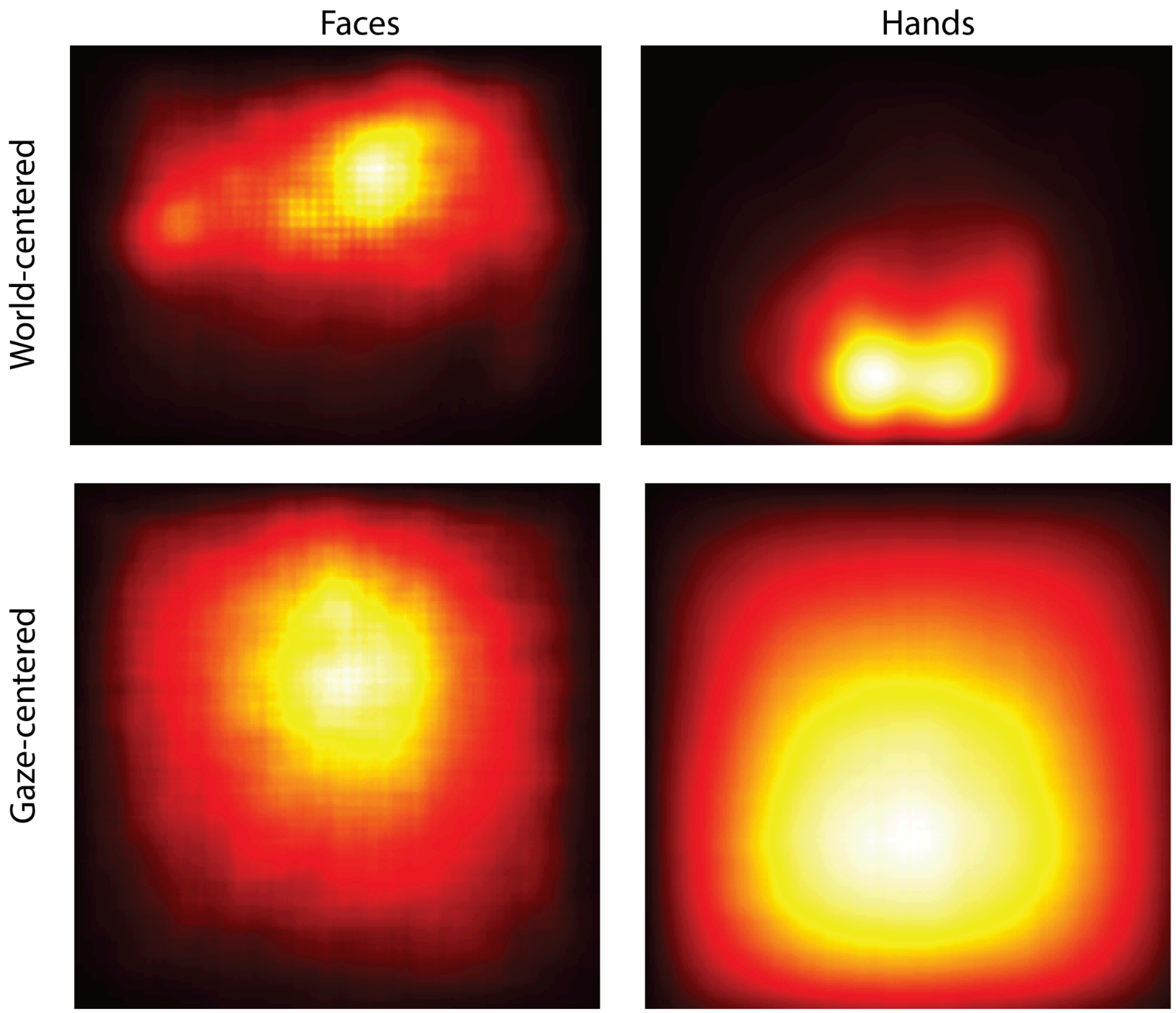

**Figure 9: (Top) Heat maps showing the location of faces and hands in 22.5 million frames of egocentric video. (Bottom) Gaze-centered heatmaps of faces and hands.**

## Access and Distribution

We are releasing most VEDB materials publicly. There are two main repositories for the materials. Databrary (https://nyu.databrary.org/volume/1612) hosts the de-identified videos from the world camera, blurred eye videos, as well as the location and task annotation files. We have also hosted the unblurred eye videos on Databrary but with limited access to protect participant privacy. Researchers who have a Databrary account will be able to view this content. Each session is named by timestamp (YYYY_MM_DD_HH_MM_SS), and is tagged with participant information (participant code, approximate age, gender, race, and ethnicity), a condition label that indicates whether the gaze tracking pipeline was successful, a basic task label and a filming location label. End users may download all content or filter according to their specific needs using the tags. The Open Science Framework (https://osf.io/2gdkb/) hosts the processed gaze files and tracking data (odometry, accelerometry, and gyroscope) because Databrary only supports video and csv file types. Both repositories name sessions by timestamp (YYYY_MM_DD_HH_MM_SS) so that corresponding files can be found. We have also released a suite of code tools for data acquisition, processing, and analysis on GitHub (https://github.com/vedb), along with introductory notebooks for using the data (https://github/com/vedb/vedb-demos). The CC-BY Creative Commons license covers all content.

## Maintenance and Sustainability

We have chosen to host the dataset on existing large-scale open science platforms (Databrary and OSF) to improve this dataset's longevity and increase its visibility in the community. We cannot guarantee the ongoing software development necessary to keep all code up-to-date. However, the CC-BY license empowers end users to use our code as a project starting point.

We are open to growth of the VEDB and encourage the community to help grow and contribute to this project to increase its utility to all.

# Author Contributions

**Greene**: conceptualization, methodology, software, investigation, data curation, writing original draft, visualization, supervision, project administration, funding acquisition.
**Balas**: conceptualization, methodology, writing review and editing, supervision, funding acquisition.
**Lescroart**: conceptualization, methodology, software, investigation, writing review and editing, supervision, funding acquisition.
**MacNeilage**: conceptualization, methodology, investigation, writing review and editing, supervision, funding acquisition.
**Hart**: methodology, investigation, data curation, project administration.
**Binaee**: methodology, software, validation, investigation, writing review and editing.
**Hausammann**: software.
**Mezile**: methodology, investigation, resources, data curation.
**Shankar**: methodology, investigation.
**Sinnott**: methodology, investigation, writing review and editing.
**Capurro:** investigation.
**Hallow**: methodology, investigation.
**Howe**: investigation.
**Josyula**: data curation, investigation.
**Li:** software, investigation.
**Mieses**: software, investigation.
**Mohamed:** investigation.
**Nudnou**: investigation.
**Parkhill:** investigation.
**Riley**: software, investigation.
**Schmidt:** investigation.
**Shinkle:** methodology, investigation.
**Si**: data curation, investigation.
**Szekely**: methodology, investigation.
**Torres:** investigation.
**Weissmann:** investigation.

# Acknowledgements

The authors would like to thank Maggie Diamond-Stanic, Jackson Skinner, and Amber Fuller for their administrative support. Thanks to Dan Gu and Robert Spellman for technical support. Annotation support was provided by Adya Agarwal, Hannah Burdick, Avery Cadorette, Olivia Cuneo, Yuleibi De Los Santos, Mavy Ho, Joey Ireland, Kaitlyn O’Shaughnessy, Colin Pierce, Samantha Simmons, and Madeline Smith.